 \documentclass[final,3p,times,twocolumn]{elsarticle}
\usepackage{amssymb}
\usepackage{hyperref}
\usepackage{graphicx}
\usepackage{subcaption}
\usepackage{lscape}
\usepackage{amsmath}
\usepackage{multirow}
\usepackage[figuresright]{rotating}
\usepackage{lineno}
\usepackage{algorithm} 
\usepackage{algpseudocode} 
\usepackage{comment}
\usepackage{lineno}

\usepackage{natbib}
\setcitestyle{authoryear,open={(},close={)},citesep={;}} 

\usepackage{calc}
\begin{document}

\begin{frontmatter}

\title{Deep Data Augmentation for Weed Recognition Enhancement: A Diffusion Probabilistic Model and Transfer Learning Based Approach}

\author[label1]{Dong Chen}
\author[label1]{Xinda Qi}
\author[label1]{Yu Zheng}
\author[label2]{Yuzhen Lu}
\author[label3]{Zhaojian Li}

\address{*Zhaojian Li (lizhaoj1@egr.msu.edu) is the corresponding author}
\address[label1]{Department of Electrical and Computer Engineering, Michigan State University, East Lansing, MI 48824, USA}
\address[label2]{Department of Agricultural and Biological Engineering, Mississippi State University, Mississippi State 39762, MS, USA \\
$^b$Department of Biosystems and Agricultural Engineering, Michigan State University, MI 48824, USA
}
\address[label3]{Department of Mechanical Engineering, Michigan State University, East Lansing, MI 48824, USA}

\begin{abstract}
Weed management plays an important role in many modern agricultural applications. Conventional weed control methods mainly rely on chemical herbicides or hand weeding, which are often cost-ineffective, environmentally unfriendly, or even posing a threat to food safety and human health. Recently, automated/robotic weeding using machine vision systems has seen increased research attention with its potential for precise and individualized weed treatment. However, dedicated, large-scale, and labeled weed image datasets are required to develop robust and effective weed identification systems but they are often difficult and expensive to obtain. To address this issue,  data augmentation approaches, such as generative adversarial networks (GANs), have been explored to generate highly realistic images for agricultural applications. Yet, despite some progresses, those approaches are often complicated to train or have difficulties to preserve fine details in images. In this paper, we present the first work of applying diffusion probabilistic models (also known as diffusion models) to generate high-quality synthetic weed images based on transfer learning. Comprehensive experimental results show that the developed approach consistently outperforms several state-of-the-art GAN models, representing the best trade-off between sample fidelity and diversity and the highest Fréchet Inception Distance ($FID$) score on a common weed dataset, CottonWeedID15 (a dataset specified to cotton production systems). In addition, the expanding dataset with synthetic weed images is able to apparently boost model performance on four deep learning (DL) models for the weed classification tasks. Furthermore, the DL models trained on the CottonWeedID15 dataset with \textit{only} 10\% of real images and 90\% of synthetic weed images achieve a testing accuracy of over 94\%, showing high-quality of the generated weed samples. The codes of this study are made publicly available at \url{https://github.com/DongChen06/DMWeeds}.
\end{abstract}


\end{frontmatter}

%% main text
\section{Introduction}
\label{sec:intro}

% weed control
Weeds are one of the major threats to crop production. Weeds compete with crops for resources, resulting in crop yield and quality losses \citep{zimdahl2007weed}.  To increase crop yields and overcome the shortage of hand weeders, chemical herbicides are being rapidly used to prevent weeds from excessive growth. It is reported that the global herbicide market grew by 39\% between 2002 and 2011 \citep{rao2017rice} and is expected to reach \$37.99 billion in 2025 \citep{ciriminna2019herbicides}. However, there is now overwhelming evidence that intensive applications of chemicals pose significant risks to humans, food safety, and environment (e.g., surface/underwater, soil, and air contamination) \citep{aktar2009impact}, as well as increasing the evolution of herbicide-resistant weeds, resulting in significant management costs. Therefore, there are urging needs to develop effective and sustainable weed management systems to reduce the environmental impact and selection pressure for weed resistance to herbicides. 

% robotic weed control and machine vision
Recently, robotic weeding combining machine vision and traditional mechanical weeding has emerged as a potential solution for site-specific and individualized weed management \citep{fennimore2019robotic}. Accurate recognition and localization of weeds is the key to achieving precise weed control and the development of efficient and robust robotic weeding systems. 
% Deep learning
To that end, various deep neural networks (DNNs) have been developed and attained great successes in various weed detection and classification tasks. Specifically, the authors in \citep{espejo2020improving} report that pretraining deep learning (DL) models on agricultural datasets are advantageous to reduce training epochs and improve model accuracy. Four DL models are fine-tuned on two datasets, Plant Seedlings Dataset and Early Crop Weeds Dataset \citep{espejo2020improving}, and classification performance improvements from 0.51\% to 1.89\% are reported.  In addition, two DL models, ResNet50 \citep{he2016deep} and Inception-v3 \citep{szegedy2016rethinking}, are tested on the DeepWeeds dataset \citep{olsen2019deepweeds} using transfer learning \citep{zhang2022review} for weed identification, achieving classification accuracies over 95\%.  In \cite{chen2022performance}, 35 state-of-the-art DL models are evaluated on the CottonWeedID15 dataset for classifying 15 common weed species in the U.S. cotton production systems, and 10 out of the 35 models obtain F1 scores over 95\%. 

% lack of labeled dataset, existing dataset
Large-scale and diverse labeled image data is essential for developing the aforementioned DL algorithms. Effective, robust, and advanced DL algorithms  for weed recognition in complex field environment require a comprehensive dataset that covers different lighting/field conditions, various weed shapes/sizes/growth stages/colors, and mixed camera capture angles \citep{zhang2022review, chen2022performance}. In contrast, insufficient or low-quality datasets will lead to poor model generalizability and overfitting issues \citep{shorten2019survey}. Recently, several  progresses have been made on developing dedicated image datasets for weed control \citep{lu2020survey}, such as DeepWeeds \citep{olsen2019deepweeds}, Early Crop Weeds Dataset \citep{espejo2020towards}, early crop dataset\citep{espejo2020towards}, CottonWeedID15 \citep{chen2022performance}, and YOLOWeeds \citep{dang2022deepcottonweeds}, just to name a few. However, collecting a large-scale and labeled image dataset is often resource intensive, time consuming, and economically costly. One sound approach to addressing this bottleneck is to develop advanced data augmentation techniques that can generate high-quality and diverse images \citep{xu2022comprehensive}. However, basic data augmentation approaches, such as geometric (e.g., flips and rotations), color transformations (e.g., Fancy PCA \citep{krizhevsky2012imagenet} and color channel swapping), tend to produce high-correlated samples and are unable to learn the variations or invariant features across the samples in the training data \citep{shorten2019survey, lu2020survey}.

% Data augmentation in agriculture
Lately, advanced data augmentation approaches, such as generative adversarial networks (GANs), have gained increased attention in the agricultural community due to their capability of generating naturalistic images \citep{LU2022107208}. 
In \cite{espejo2021combining}, Deep Convolutional GAN (DCGAN) \citep{radford2015unsupervised} combined with transfer learning is adopted to generate new weed images for weed identification tasks. The authors then train the Xception network \citep{chollet2017xception} with the synthetic images on the Early Crop Weed dataset \citep{espejo2020towards} (contains 202 tomato and 130 black nightshade images at early growth stages) and obtain the testing accuracy of 99.07\%. Conditional Generative Adversarial Network (C-GAN) \citep{mirza2014conditional} is adopted in \citep{abbas2021tomato} to generate synthetic tomato plant leaves to enhance the performance of plant disease classification. The DenseNet121 model \citep{huang2017densely} is trained on synthetic and real images using transfer learning on PlantVillage dataset \citep{hughes2015open} to classify the tomato leaves images into ten categories of diseases, yielding 1-4\% improvement in classification accuracy compared to training on the original data without image augmentation. The readers are referred to \citep{LU2022107208} for more recent advances of GANs in agricultural applications. While impressive progresses have been made, GANs often suffer from training instability and model collapse issues \citep{wgan2017, creswell2018generative, mescheder2017numerics}, and they could fail to capture a wide data distribution \citep{zhao2018bias}, which make GANs difficult to be scaled and applied to new domains.

% Diffusion model
On the other hand, diffusion probabilistic models (also known as diffusion models) have quickly gained popularity in producing high-quality images \citep{ho2020denoising, song2020denoising, dhariwal2021diffusion}. Diffusion models, inspired by non-equilibrium thermodynamics \citep{jarzynski1997equilibrium, sohl2015deep}, progressively add noise to data and then construct the desired data samples by a Markov chain from white noise \citep{song2020denoising}. Recent researches show that diffusion models are able to produce  high-quality synthetic images, surpassing GANs on several different tasks \citep{dhariwal2021diffusion}, and there is significant interest in validating diffusion models in different image generation tasks, such as video generation \citep{ho2022video}, medical image generation \citep{ozbey2022unsupervised}, and image-to-image translation \citep{saharia2021palette}. However, the potential of diffusion methods in agricultural image generation remains largely unexplored, partly owing to the substantial computational burden of image sampling in regular diffusion models. In this paper, we presented the \textit{first} results on image generation for weed recognition tasks using a classifier-guided diffusion model (ADM-G, \citep{dhariwal2021diffusion}) based on a 2D U-Net \citep{ronneberger2015u} diffusion model architecture.  
A common weed dataset, CottonWeedID15 \citep{chen2022performance}, is used and evaluated on the image generation and classification tasks.  The main contributions and the technical advancements of this paper are highlighted as follows.
% contributions
\begin{enumerate}
    \item We present the first study on applying diffusion models through transfer learning to generate high-quality images for weed recognition tasks based on a popular weed dataset, CottonWeedID15.
    \item A post-processing approach based on realism score is developed and employed to automatically remove low-quality samples and underrepresented samples after the training to improve the quality of the generated samples. 
    \item Four DL models are evaluated on the expanding dataset with synthetic images through transfer learning, showing significant performance improvement on the weed classification accuracy.
    \item We conduct comprehensive experiments, and the results show that the proposed approach consistently outperforms several state-of-the-art GANs in terms of sample quality and diversity. The codes of this study are open-sourced at: \url{https://github.com/DongChen06/DMWeeds}.
\end{enumerate}

The rest of the paper is organized as follows. Section \ref{sec:data_collection} presents the dataset and technical details used in this study. Experimental results and analysis are presented in Section~\ref{sec:results}. Finally, concluding remarks and potential future works are provided in Section ~\ref{sec:conclu}.

\section{Materials and Methods}
\label{sec:data_collection}

\subsection{CottonWeedID15 dataset}\label{sec:cottonweedid15}
To demonstrate the effectiveness of the proposed approach, we evaluate the developed approach on a common weed dataset, CottonWeedID15 \citep{chen2022performance}, which is an open-source weed dataset for weed recognition tasks collected under natural light conditions at various weed growth stages in the southern United States, containing 5,187 images of 15 common weed classes in the cotton production systems. The dataset has unbalanced classes and the images are saved in ``jpg" format at varied resolutions. It includes five major classes,  Morningglory, Carpetweed, Palmer Amaranth, Waterhemp, and Purslane, with image numbers ranging from 450 to 1,115, while the minority classes, such as Swinecress and Spurred Anoda, have less than 100 images. To mitigate the effect of unbalanced classes and facilitate the model training, we only included the top 10 weed classes (4,685 images) in this study and each weed class contains at least 200 images. The dataset was randomly partitioned into  training (90\%) and testing (10\%) sets, in which the training subset was used to train the diffusion models (Section~\ref{sec:dm}) as well as GAN baselines (Section~\ref{sec:baselines}) whereas the testing set was hold out to test the performance of the DL weed classifiers (Section~\ref{sec:classifiers}). To save computation resources, all weed images were resized to $256 \times 256$. The sample images from CottonWeedID15 dataset are shown in the top three rows in Figure~\ref{fig:sample_cottonweeds}.

\begin{figure*}[!ht]
  \centering
  \includegraphics[width=0.7\textwidth]{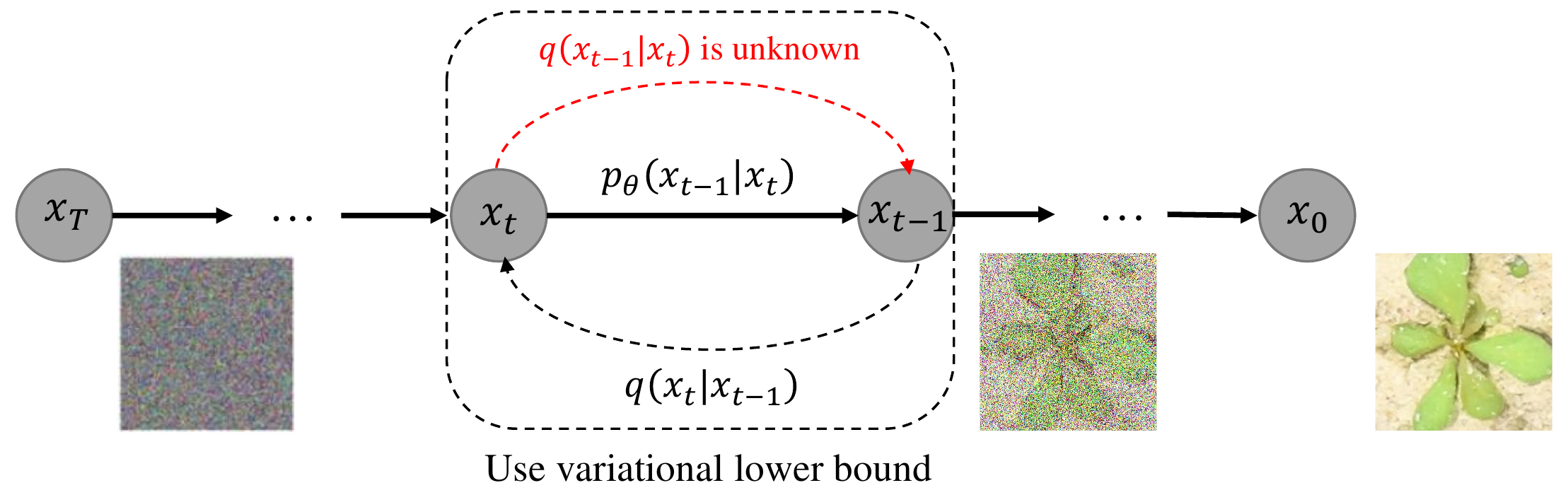}
  \caption{Illustration of the Markov chain modeling on the forward/inverse diffusion process for generating a sample by slowly adding/removing noises.}
  \label{fig:diffusion_model}
\end{figure*}

\subsection{Diffusion models}
% introduce of what is diffusion
The concept of diffusion is widely used in many fields, such as physics \citep{libbrecht2005physics}, thermodynamics \citep{qian2015thermodynamics}, statistics \citep{florens1989approximate}, and finance \citep{eraker2001mcmc}, referred to as a process transferring from one distribution to another distribution. For example, in thermodynamics, diffusion process often refers to the movement of molecules from a region of high concentration to another region of lower concentration \citep{qian2015thermodynamics}. In \cite{sohl2015deep}, the authors, inspired by non-equilibrium statistical physics \citep{jarzynski1997equilibrium},  formulate the diffusion process as a Markov chain and gradually convert data distribution into another known distribution (e.g., Gaussian) through an iterative forward diffusion process, then construct new data from the random noise via a reverse diffusion process, the details of which will be discussed next. %The idea of using diffusion models for weed image generation will be introduced in the following subsections.

\subsubsection{Forward diffusion process}
Forward diffusion process \citep{song2020denoising} is the process that transfers a complex data distribution $p_{data}$ to a simple, known distribution $p_{prior}$, e.g., $p_{prior} \sim \mathcal{N}(0, I)$. Specifically, it first samples a data point from the real data distribution $x_0 \sim q(x)$ and gradually adds a small amount of noise to the sample in $T$ steps,  producing a sequence of noisy samples $x_1, x_2, \ldots, x_T$. The probability of transforming from sample $x_{t-1}$ to $x_t$ is modeled by $q(x_t|x_{t-1})$, and the objective is that when $T$ approaches $+ \infty$ , $x_T$ will approach the known distribution $p_{prior}$. In \cite{sohl2015deep}, the authors show that $q(x_t|x_{t-1})$ follows a Gaussian distribution with the following form:
\begin{equation} \label{eqn:q}
    q(x_t|x_{t-1}) = \mathcal{N}(x_t; \sqrt{1-\beta_t}x_{t-1}, \beta_t I),
\end{equation}
where $\beta_t \in (0, 1)$ is the diffusion rate at step $t$ and it can be held constant or learned online \citep{ho2020denoising}.
Applying reparameterization tricks \citep{kingma2015variational}, the above process is shown to be the same as $x_t = \sqrt{1-\beta_t} x_{t-1} + \beta_t z_{t-1}$, i.e., combing $x_{t-1}$ and standard normal distribution $z_{t-1} \sim \mathcal{N}(0, I)$, where $\beta_t$ controls the amount of perturbations. Let $a_t = 1 - \beta_t$ , $\bar{a}_t = \prod_{i=1}^{T} a_i$, $z_{t-1}, z_{t-2}, \ldots, \sim \mathcal{N}(0, I)$, and merged Gaussian distribution $\bar{z}_{t-2}, \bar{z}_{t-3}, \ldots, \sim \mathcal{N}(0, I)$, then we can rewrite $x_t$ in terms of $x_0$ as follows:
\begin{equation} \label{eq:xt}
\begin{split}
x_t & = \sqrt{a_t} x_{t-1} + \sqrt{1-a_t} z_{t-1} \\
 & = \sqrt{a_t a_{t-1}}x_{t-2} + \sqrt{a_t (1-a_{t-1})}z_{t-2} + \sqrt{1-a_t}z_{t-1} \\
 & = \sqrt{a_t a_{t-1}}x_{t-2} + \sqrt{1-a_t a_{t-1}}\bar{z}_{t-2} \\
 & = \ldots \\
 & = \sqrt{\bar{a}_t}x_0 + \sqrt{1-\bar{a}_t} \bar{z},
\end{split}
\end{equation}
where $\bar{z}$ is a merged Gaussian distribution. From which it follows that we can represent $q(x_t|x_{t-1})$ as:
\begin{equation} \label{eqn:q}
    q(x_t|x_0) = \mathcal{N}(x_t; \sqrt{\bar{a}_t}x_0, (1-\bar{a}_t) I).
\end{equation}
Since $\beta_t \in (0, 1)$, we have  $a_t \in (0, 1)$ and $\bar{a}_t \rightarrow 0$ when $t \rightarrow \infty$, thus we have $q(x_T) = \mathcal{N}(0, I)$. Usually, when $T$ is large enough, e.g., $T\ge 4000$, $q(x_t)$ is close to  $\mathcal{N}(0, I)$.

\subsubsection{Reverse diffusion process}
The target of the reverse diffusion process is to reconstruct a synthetic data sample from a random sample from the known distribution $p_{prior}$, which is the reverse process of the above-mentioned forward diffusion process, i.e., $x_T \sim \mathcal{N}(0, I) \rightarrow x_0 \sim p_{data}$. The reverse process can be represented as the transformation $q(x_{t-1} | x_t),\, t \in \{T, T-1, T-2, \ldots, 0 \}$. In \cite{feller1949theory}, the authors prove that the transformation in the forward process $q( x_t | x_{t-1})$ and reverse process $q(x_{t-1} | x_t)$ often share the same format of distributions. However, it is generally difficult to find the parameters of $q(x_{t-1} | x_t)$. Instead, one can approximate $q(x_{t-1} | x_t)$ with a function approximator $p_{\theta}(x_{t-1} | x_t)$ as follows:
\begin{equation}\label{eq:p_x}
    p_{\theta}(x_{t-1} | x_t) = \mathcal{N} \left( x_{t-1}; \mu_{\theta}(x_t, t), \Sigma_{\theta} (x_t, t) \right),
\end{equation}
where $\theta$ is the learnable parameter vector in the mean function $\mu_{\theta}(x_t, t)$ and the standard deviation function $\Sigma_{\theta} (x_t, t))$ of the Gaussian distribution. As a result, the distribution of the generated data samples can be represented as:
\begin{equation}
    p_{\theta}(x_{0:T}) = p(x_T) \prod_{t=1}^{T}p_{\theta}(x_{t-1} | x_t).
\end{equation}
Therefore, with the learned parameters $\theta$ and the resultant $\mu_{\theta}(x_t, t)$ and $\Sigma_{\theta} (x_t, t)$, one can generate a data sample from a random sample $x_T \sim \mathcal{N}(0, I)$ through the reverse diffusion process. The aforementioned procedures are illustrated in Figure~\ref{fig:diffusion_model}. In summary, the forward diffusion process is to add noise to the data sample, while the reverse diffusion process is to remove the noise and reconstruct new data samples.

\subsubsection{Training objectives}
The parameterized functions $\mu_{\theta}(x_t, t)$ and $\Sigma_{\theta} (x_t, t)$ are often characterized by DL models, which can be trained by minimizing the cross entropy  between $q(x_0)$ and $p_{\theta}(x_0)$, i.e., $-E_{q(x_0)} \log p_{\theta}(x_0)$.  Similar to  Variational Autoencoder (VAE, \cite{kingma2013auto}), minimizing the negative log-likelihood can be achieved by minimizing the variational lower bound loss as follows:
\begin{equation}\label{eq:loss_vlb}
   L_{VLB} = E_{q(x_{0:T})} \left[\log \frac{q(x_{1:T}|x_0)}{p_{\theta}(x_{0:T})} \right] \geq -E_{q(x_0)} \log p_{\theta}(x_0),
\end{equation}
where $q(x_{1:T}|x_0) = \prod_{t=1}^{T} q(x_t|x_{t-1})$. According to \citep{weng2021diffusion, sohl2015deep}, the above objective function can be further rewritten as: 
\begin{equation}\label{eq:loss}
\begin{split}
L_{VLB} & = E_q \left[ -\log p_{\theta}(x_0|x_1) \right]  \\
 &\qquad + \sum_{t=2}^{T} D_{KL} \Bigl( q(x_{t-1}|x_t, x_0) ~\|~ p_{\theta}(x_{t-1}|x_t) \Bigl)  \\
 &\qquad + D_{KL} \Bigl( q(x_T|x_0)~\|~p_{\theta}(x_T) \Bigl) \\
 & = L_0 + \sum_{t=2}^{T} L_{t-1} + L_T.
\end{split}
\end{equation}
Here $L_T$ is a constant because $x_0$ is from a known data distribution and $x_T$ is a Gaussian noise. Note that since $p_{\theta}(x_0|x_1) = \mathcal{N}(\mu_{\theta}(x_1, 1), \delta^2_1 I) $, $L_0$ can be viewed as the entropy of the multivariate Gaussian distribution and it is a constant only depending on $\delta^2_1 I$. 

The loss term $L_t, t \in [1, 2, 3, \ldots, T-1 ]$ can be parameterized as:
\begin{equation}\label{eq:L_t}
  \resizebox{.98\hsize}{!}{$L_t = E_{x_0, z_t} \left[\frac{\beta^2_t}{2\alpha_t(1 - \bar{\alpha}) \delta^2_t}~\big\| z_t - z_{\theta}(\sqrt{\bar{\alpha}_t}x_0 + \sqrt{1-\bar{\alpha}_t }z_t, t) \big\|^2 \right] + C $},
\end{equation}
where $C$ is a constant not depending on $\theta$. In DDPM \citep{ho2020denoising}, the authors argue that the diffusion models perform better with a simplified objective $L_t$ that ignores the weighting term $\frac{\beta^2_t}{2\alpha_t(1 - \bar{\alpha}) \delta^2_t}$ as follows:
\begin{equation}\label{eq:L_ts}
  \resizebox{.98\hsize}{!}{$L_{simple}(\theta) = E_{x_0, z_t} \left[ \big\| z_t - z_{\theta}(\sqrt{\bar{\alpha}_t}x_0 + \sqrt{1-\bar{\alpha}_t }z_t, t) \big\|^2 \right] + C$. }
\end{equation}

\subsubsection{Conditional generation}
To generate samples conditioning on class labels $y$ with diffusion models, it is common to incorporate class information $y$ to the diffusion process, i.e., replacing $p(x_{t-1} | x_t)$ with $p(x_{t-1} | x_t, y)$. However, in ADM-G \citep{dhariwal2021diffusion},  the authors argue that naive class-conditional diffusion models often produce low-quality and incoherent samples. To overcome this issue, the authors train a classifier $p(y | x_t)$ that predicts the label for the noise-injected image $x_t$ and perturbs the mean of the denoising step $p(x_{t-1}|x_t)$ by the scaled gradient of the classifier $\nabla \log (p(y|x_t))$, which guides the diffusion sampling process towards the conditioned information $y$ (e.g., a target class label) by altering the noise prediction \citep{weng2021diffusion, dhariwal2021diffusion}.
The denoising function Eq.~\ref{eq:p_x} can be rewritten as:
\begin{equation}\label{eq:adm}
  p ( x_{t-1} | x_t, y ) = \mathcal{N} \left(\mu_{\theta} + s \sum \nabla \log (p(y|x_t)), \Sigma_{\theta} \right),
\end{equation}
where $s$ is the guidance scale (or classifier scale), and a larger classifier scale will help the model focus more on the  classifier and generate higher-fidelity but less diverse samples \citep{dhariwal2021diffusion}. In this study, we  present the first study on applying the classifier-guided diffusion model (ADM-G) to generate high-quality weed images through transfer learning for weed classification tasks, which will be detailed in the following subsections.

\subsection{DL models for weed classification} 
To evaluate the quality of synthetic weed images generated by the diffusion models, we show that training the weed classifiers with the augmented images can boost performance improvement for weed recognition tasks. In this study, four state-of-the-art DL models were selected for classifying the weed images: ResNet50 \citep{he2016deep}, Inception-v3 \citep{szegedy2016rethinking}, VGG16 \citep{ding2021repvgg}, and DenseNet121 \citep{huang2017densely}. These models are widely adopted and evaluated for classifying weed images \citep{olsen2019deepweeds, ahmad2021performance, chen2022performance}. Conventional cross-entropy loss is adopted to train the DL models as:
\begin{equation}\label{eq:ce}
  L_{CE} = -\sum_{c=1}^{M} y_{i, c} \log (p_{i, c}),
\end{equation}
where $M$ is the number of weed classes; $y_{i,c}$ is a binary indicator (0 or 1) and it equals to 1 if the class label $c$ is the correct classification for  input $i$; and $p_{i,c}$ is the Softmax probability \citep{goodfellow2016deep} for the $c$th class.

\subsection{Transfer learning} \label{sec:tl}
In this paper, we adopt transfer learning \citep{zhuang2020comprehensive} to speed up the learning and improve the performance of the diffusion models as well as the weed classifiers. Instead of training diffusion models and weed classifiers from scratch, we train the models by fine-tuning the models pre-trained on ImageNet dataset \citep{russakovsky2015imagenet}. 

% Transfer learning for diffusion models
To apply transfer learning to speed up the training of diffusion models, we adopt pre-trained models\footnote{\url{https://github.com/openai/guided-diffusion}} as the starting point, and then fine-tune the parameters of the diffusion models on the CottonWeedID15 dataset. As in ADM-G,  a downsampling trunk (encoder part) of the U-Net model is chosen as the classifier architecture with an attention pool at the $8 \times 8$ layer to produce the final output.  Therefore, the size of the output layer is changed from 1,000 (1000 object classes in ImageNet)  to 10 (number of classes in CottonWeedID15).

% Transfer learning for weed classifiers
To apply transfer learning for training weed classifiers, DL models are first initialized on the pre-trained models\footnote{\url{https://pytorch.org/vision/stable/models.html}} on ImageNet, and  all the model parameters are then updated according to the weed classification tasks. In addition, DL models are modified to be consistent with the number of weed classes. Specifically, in ResNet50 \citep{he2016deep}, VGG16 \citep{simonyan2014very}, Inception-v3 \citep{szegedy2016rethinking} and DenseNet121 \citep{huang2017densely}, the final layers are reshaped to have the same number of outputs as the number of weed classes in the CottonWeedID15 (i.e., 10 classes) dataset. However, the original implementation of Inception-v3 only works with input image sizes larger than $299 \times 299$ if the auxiliary network is used, but the image resolution used in this study is $256 \times 256$. Thus, the auxiliary network in Inception-v3 is not activated during the training to adapt to smaller input sizes, i.e., $256 \times 256$.

\subsection{Performance evaluation metrics}

\subsubsection{Evaluation metrics for diffusion models}
Evaluating the quality of synthetic images generated by GANs or diffusion models is a daunting task and an open problem \citep{goodfellow2016nips, dhariwal2021diffusion}. Typically, two properties, \textit{image quality} (also named image fidelity) and \textit{image diversity}, of the generated images, are evaluated, where image fidelity indicates whether the synthetic images look like a specific object whereas image diversity measures whether diverse images (or a wide distribution of images) are generated.
In this study, the generated images are evaluated in terms of Fréchet Inception Distance ($FID$) \citep{heusel2017gans},  Inception Score ($IS$) \citep{salimans2016improved}, Improved Precision and Recall ($Prc$ and $Rec$) \citep{kynkaanniemi2019improved},  which are described below. Specifically,  we use $FID$ as the overall sample quality measure. Sample fidelity is measured using $Prc$ and $IS$, while sample diversity or distribution coverage are evaluated with $Rec$.

\textbf{Fréchet Inception Distance ($FID$)} was first proposed by \citep{heusel2017gans} to measure the quality of generated images by calculating the distance in the feature vector space between real and generated images. Lower $FID$ indicates higher image quality and that the images are more similar and likely to come from close data distributions. $FID$ extracts the feature representations from the last pooling layer before the output of the DL model (i.e., Inception-v3 \citep{szegedy2016rethinking} model). Then the feature vectors are represented as a multivariate Gaussian vector by calculating the mean and covariance of the features. The distance between the distributions of the real and generated images is measured using the Frechet distance \citep{frechet1957distance} (also known as Wasserstein-2 distance). The use of feature extraction from the Inception-v3 model and that it is measured with the Frechet distance give the score name of ``Frechet Inception Distance”.

\textbf{Inception Score ($IS$)} was proposed by \citep{salimans2016improved} to measure the quality of generated images by GANs. Specifically, generated images are evaluated by a pre-trained DL model (e.g., Inception-v3 \citep{szegedy2016rethinking}) on the image classification task, contributing to the name ``Inception" Score). To compute $IS$, the DL model will predict the classes for all the generated images to get the conditional label distribution $p(y|x)$. To have high image quality, it is desired to have low entropy of the conditional label distribution $p(y|x)$. In addition, we expect the model to generate diverse images, so the marginal $\int p(y|x = G(z))dz$ should have high entropy. In summary, $IS$ is computed by combing the above two objectives as follows:
\begin{equation} \label{is}
    IS = exp \left( E_x KL(p(y|x)~\|~p(y)) \right),
\end{equation}
where $exp$ represents exponentiation operation and $KL$ denotes the Kullback–Leibler (KL) divergence. Higher $IS$ represents a higher quality of the generated images. In this paper, we also adopt the Inception-v3 \citep{szegedy2016rethinking} to predict the image classes but with the pre-trained model on DeepWeedID15 dataset \citep{chen2022performance}.

\textbf{Improved Precision and Recall}  ($Prc$ and $Rec$) was proposed in \citep{kynkaanniemi2019improved} to measure the sample fidelity ($Prc$) and diversity ($Rec$). To measure $Prc$ and $Rec$, a set of real and generated image samples are drawn from the real distribution ($P_r$) and the generated  distribution ($P_g$), i.e., $X_r \sim P_r$ and $X_g \sim P_g$. Then the sampled images are encoded into a high-dimensional feature space via a pre-trained VGG16 network \citep{simonyan2014very}  as $\phi_r$ and $\phi_g$, respectively, where the set of extracted feature vectors are represented as $\Phi_r$ and $\Phi_g$. To determine whether a given sample $\phi$ belongs to a distribution $\Phi$, a binary function is defined as:
\begin{equation}
    f(\phi, \Phi) = 
    \begin{cases}
          1, &\text{if } ||\phi - \phi'||_2 \leq ||\phi' - NN_k(\phi', \Phi)||_2, \\
          0, &  \text{otherwise},
    \end{cases}
\end{equation}
where $NN_k(\phi', \Phi)$ returns the $k$th nearest feature vector of $\phi'$  from $\Phi$ by calculating pairwise Euclidean distances \citep{kynkaanniemi2019improved}; and $k$ is a hyperparameter controlling the size of the neighborhood, and higher values of k increase the $Prc$ and $Rec$ estimates but come with more computation cost. In this study, we find $k=3$ is a robust choice.
If $f(\phi, \Phi) =1$ for at least one $\phi' \in \Phi$, it means that the sample $\phi$ belongs to the data distribution of $\Phi$. Then, $Prc$ and $Rec$ are defined as:
\begin{equation} \label{ip}
    Prc(\Phi_r, \Phi_g) = \frac{1}{|\Phi_g|} \sum_{\phi_g \in \Phi_g} f(\phi_g, \Phi_r),
\end{equation}

\begin{equation} \label{ir}
    Rec(\Phi_r, \Phi_g) = \frac{1}{|\Phi_r|} \sum_{\phi_r \in \Phi_r} f(\phi_r, \Phi_g),
\end{equation}
where $Prc(\Phi_r, \Phi_g)$ measures the fraction of generated samples that fall into the distribution of real images, while $Rec(\Phi_r, \Phi_g)$ measures the fraction of real samples that fall into the distribution of generated images.

\subsubsection{Evaluation metrics for weed classifiers}
To evaluate the performance of DL models on the weed classification tasks, common metrics for image classification, such as precision, recall, and Top-1 accuracy, are employed.

For classification tasks, each prediction can be classified into the following four categories: true positive ($T_P$), false positive ($F_P$), false negative ($F_N$), and true negative ($T_N$), by comparing the prediction result with the true label. \textbf{Precision} measures the fraction of positive predictions that are correctly predicted and is defined as follows:
\begin{equation} \label{eq:precision}
    \text{Precision} = \frac{T_P}{T_P + F_P}.
\end{equation}

\textbf{Recall} is the fraction of the positive samples that are successfully predicted and defined as:
\begin{equation} \label{eq:recall}
    \text{Recall} = \frac{T_P}{T_P + F_N}.
\end{equation}

\textbf{Top-1 accuracy} measures the proportion of correct predictions and is defined as:
\begin{equation} \label{eq:acc}
   \text{Top-1 accuracy} = \frac{T_P + T_N}{T_P + F_P + F_N + T_N}.
\end{equation}

\subsection{Realism score-based data cleanup}\label{sec:realism}
% motivation for data cleanup
Despite great progresses in generating synthetic samples, generative models often produce low-quality and underrepresented samples \citep{mo2019mining, lee2021self}, due to model collapse to minor data instances in the training dataset. In this study, we develop a post-processing data selection approach to remove those low-quality samples after the training based on a realism score-based approach \citep{kynkaanniemi2019improved}. Realism score ($R$) estimates how close an individual synthetic sample is to the real data and it is defined as follows: 
\begin{equation} \label{eq:r}
    R(\phi_g, \Phi_r) = \max_{\phi_r} \Biggl\{ \frac{\| \phi_r - NN_k (\phi_r, \Phi_r)\|_2}{\| \phi_g - \phi_r\|_2} \Biggl\},
\end{equation}
where $\phi_g$ and $\phi_r$ are the extracted feature vectors of the synthetic image and real image, respectively. If $R \geq 1$, the feature vector $\phi_g$ is located inside the neighboring hypersphere of at least one $\phi_r$. To filter out the low-quality images, we compute the realism score $R$ for each individual generated image, and the generated images with $R < 1$ are removed and not used to train the DL models for weed classification.

\subsection{Experimental settings}
In this study, we adopt the diffusion model design as in ADM-G \citep{ho2020denoising, dhariwal2021diffusion} and a 2D U-Net architecture \citep{ronneberger2015u} is adopted as the diffusion model architecture. In addition, other tricks, such as multi-resolution attention layers, two residual blocks per resolution, and BigGAN-like up/downsampling, are adopted to further boost the model performance. We train the diffusion models for 300,000 steps and the class classifier for 200,000 steps with the image resolution at $256 \times 256$. The pre-trained models \footnote{\url{https://github.com/openai/guided-diffusion}} for the diffusion models and class classifiers on ImageNet are utilized for transfer learning (Section~\ref{sec:tl}), which are then finetuned on the CottonWeedID15 dataset.

To train DL models for weed classification,  conventional data augmentation methods, such as random rotation for 360 degrees and random horizontal flip are applied to enhance the model training.  We train all DL models for 70 epochs with the SGD (stochastic gradient descent) optimizer and a momentum of 0.9. To achieve a reliable performance evaluation, DL models are trained with 5 replications, with different random seeds that are shared by all the models, and the mean accuracy and standard deviation on the testing set are calculated for performance evaluation according to the Monte Carlo cross-validation approach \citep{raschka2018model}.
The learning rate and batch size are set as 0.001 and 32, respectively. All experiments are implemented with the deep learning library, PyTorch (version 1.10) with Torchvision (version 0.10.0) \cite{paszke2019pytorch}. To speed up the training, a multiprocessing package\footnote{\url{https://docs.python.org/3/library/multiprocessing.html}} with 8 CPU cores is employed during the training. The experiments are performed on an Ubuntu 20.04 server with an AMD 3990X 64-Core CPU and a GeForce RTX 3090Ti GPU (24 GB GDDR6X memory).

\section{Results}
\label{sec:results}

\subsection{Evaluation of diffusion models}\label{sec:dm}
Table~\ref{tab:classifier_scales} shows the performance of the diffusion models with different classifier scales. Generally, larger classifier scales help the model generate higher fidelity (i.e., higher $Prc$ and $IS$) but less diverse samples (i.e., lower $Rec$). That is because larger classifier scale typically makes the model focus more on the guided classifier. For example, $IS$ increases from 6.37 to 7.08 as we increase the classifier scale from 0.5 to 5.0. 
In the following experiments, we chose classifier scale $s = 1.0$ since it gives the best $FID$ of 24.14. The sample images generated with different classifier scales are shown in Appendix~\ref{fig:classifier_scales}.

\begin{table*}[!ht]
\centering
\renewcommand{\arraystretch}{1.0}
\resizebox{0.65\textwidth}{!}{
\begin{tabular}{llcccccc}
\hline
\multirow{2}{*}{Dataset}       & \multirow{2}{*}{Metrics} & \multicolumn{4}{c}{Classifier scales}                                                       \\ \cline{3-8} 
                                &                          & \multicolumn{1}{c}{$s=0.5$} & \multicolumn{1}{c}{$s=1.0$} & \multicolumn{1}{c}{$s=2.0$} & \multicolumn{1}{c}{$s=3.0$}  & \multicolumn{1}{c}{$s=4.0$} & $s=5.0$ \\ \hline \hline
\multirow{4}{*}{CottonWeedID15} & FID                      & \multicolumn{1}{c}{25.66}      & \multicolumn{1}{c}{\textbf{24.14}}      & \multicolumn{1}{c}{25.65}      &    \multicolumn{1}{c}{26.75}  &    \multicolumn{1}{c}{27.26}    &   28.69 \\ 
                                & IS                       & \multicolumn{1}{c}{6.36}      & \multicolumn{1}{c}{6.70}      & \multicolumn{1}{c}{6.91}      &   \multicolumn{1}{c}{6.92}    &    \multicolumn{1}{c}{6.99}  &   \textbf{7.08}  \\ 
                                & Prc                      & \multicolumn{1}{c}{0.7065}      & \multicolumn{1}{c}{0.6805}      & \multicolumn{1}{c}{0.7120}      &    \multicolumn{1}{c}{0.7025}   &    \multicolumn{1}{c}{0.7065}   &  \textbf{0.7145}  \\ 
                                & Rec                      & \multicolumn{1}{c}{0.7095}      & \multicolumn{1}{c}{\textbf{0.8035}}      & \multicolumn{1}{c}{0.6780}      &   \multicolumn{1}{c}{0.7470}   &    \multicolumn{1}{c}{0.7225}   &  0.6895   \\ \hline
\end{tabular}}
\caption{Performance evaluation of the proposed diffusion models with different classifier scale factors. Best values are in bold.}
\label{tab:classifier_scales}
\end{table*}

\begin{figure*}[!ht]
  \centering
  \includegraphics[width=0.95\textwidth]{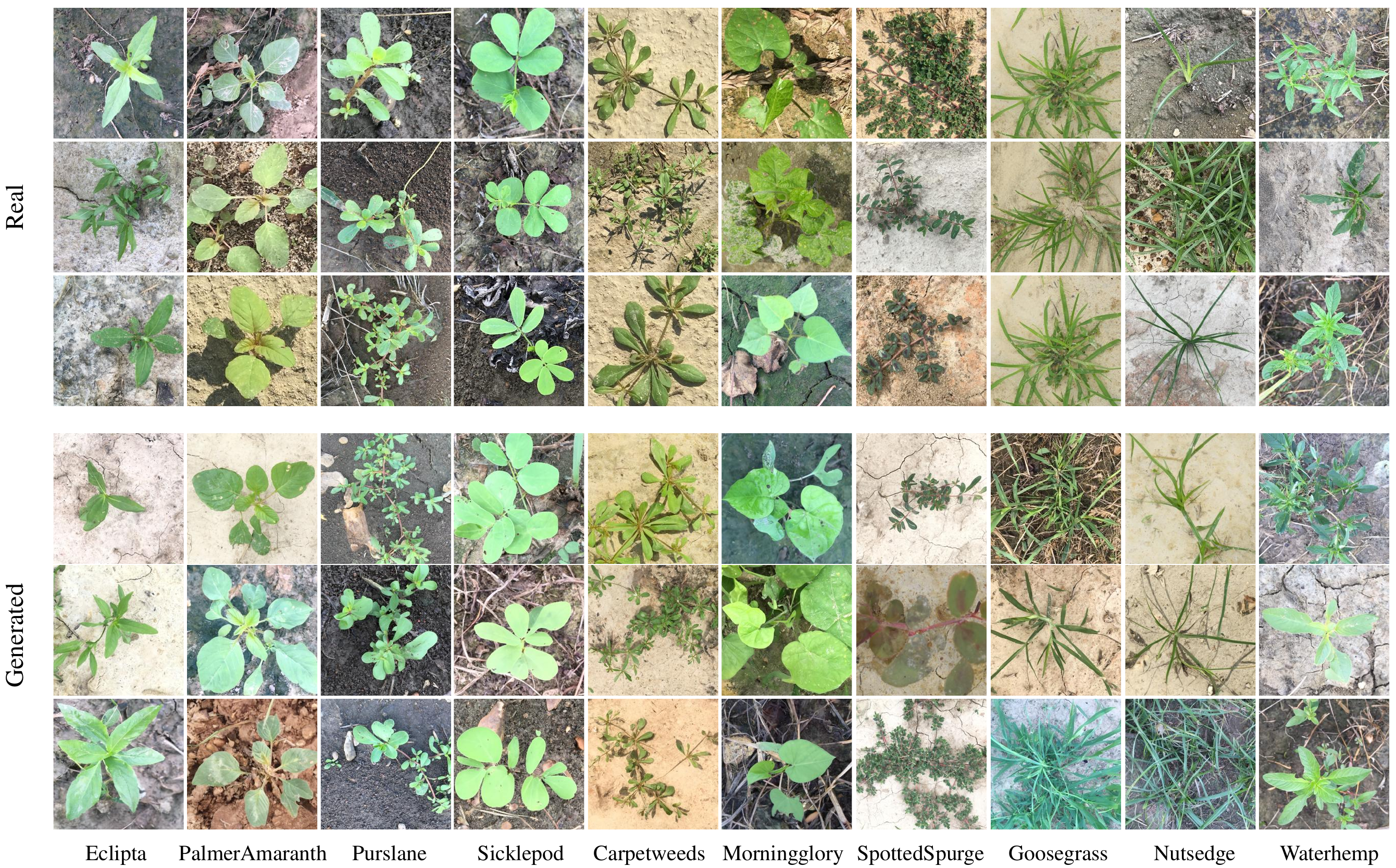}
  \caption{Real (upper panel) and generated (lower panel) weed samples from CottonWeedID15 model. Each column represents one weed class.}
  \label{fig:sample_cottonweeds}
\end{figure*}

Figure~\ref{fig:sample_cottonweeds} shows the generated and the real samples from the CottonWeedID15 dataset. It is obvious that the proposed approach is able to generate realistic and diverse images. The reason is that the diffusion model produces diverse samples by learning variant features across all the samples in the training sets and learns the data distribution, which is evident by the different backgrounds observed in generated images.

\begin{figure*}[!ht]
  \centering
  \includegraphics[width=0.95\textwidth]{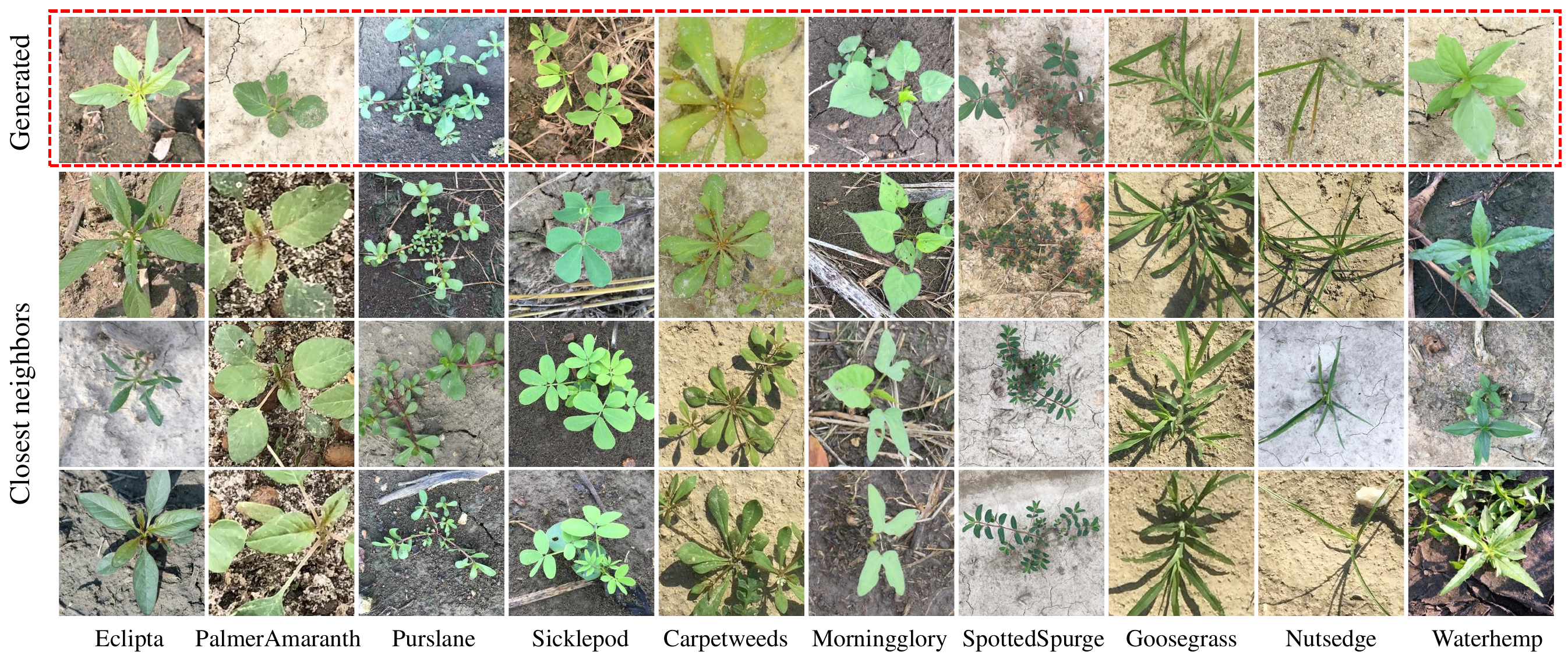}
  \caption{Nearest neighbor samples in the CottonWeedID15 dataset for each generated weed image. Each column represents one weed class, and the top row shows synthetic samples whereas the remaining rows are the top 3 nearest neighbors from the CottonWeedID15 dataset.}
  \label{fig:neighbors_cottonweeds}
\end{figure*}

To show that the synthetic samples are not generated by recalling existing images from the training datasets, we searched for the nearest neighbors in the training dataset for the generated samples by comparing the distance in the feature space of VGG16 network \citep{simonyan2014very}  between the synthetic and real images. Figure~\ref{fig:neighbors_cottonweeds} shows the Top-3 nearest neighbors from CottonWeedID15 dataset for the generated samples. It is clear that the proposed approach indeed generated unique samples that are not present in the training dataset.

\subsection{Comparison with state-of-the-art benchmarks}\label{sec:baselines}
In this subsection, we compare the proposed method with
several state-of-the-art GANs: BigGAN \citep{brock2018large}, StyleGAN2 \citep{karras2020analyzing}, and StyleGAN3 \citep{karras2021alias}. All GAN models are fine-tuned on the pre-trained models on ImageNet, and are then trained on the CottonWeedID15 dataset for 300,000 steps. The two variants of the diffusion model (ADM-G), ADM and ADM-DDIM, stand for the diffusion model without the class classifier and the diffusion model sampled with DDIM \citep{song2020denoising} (allows for fewer sampling steps, e.g., 25 steps in our experiment), respectively. Table~\ref{tab:benchmarks} shows the evaluation results on the CottonWeedID15 dataset. Overall, diffusion model-based approaches outperform the state-of-the-art GANs. In addition, the classifer-guided diffusion model (ADM-G) achieves the highest $FID=24.14$ and produces much more natural weed images (see Figure~\ref{fig:sample_cottonweeds}). 
ADM-G achieves slightly lower $Rec$ than ADM but higher $FID$, $IS$, and $Prc$ since class classifier tends to guide the diffusion models to generate high-quality samples by sacrificing some diversity. This is consistent with observations in other works, see e.g., \citep{dhariwal2021diffusion}. ADM-DDIM performs worse than ADM and ADM-G due to fewer sampling steps.
BigGAN fails to generate high quality images due to the grid effects \citep{odena2016deconvolution, zhang2019gan} (see generated sample images in Appendix~\ref{fig:sample_cottonweeds_biggan}). StyleGAN2 is able to generate reasonably good images with the highest $IS=7.39$ and $Prc=0.6850$, but it generates less diverse samples with a relatively lower $Rec=0.4430$ (see  generated sample images in Appendix~\ref{fig:sample_cottonweeds_stylegan2}). For example, StyleGAN2 fails to generate diverse samples for Goosegrass weed and all generated samples are identical. StyleGAN3 ($FID=26.28$) achieves better trade-off between image quality and diversity than StyleGAN2 ($FID=43.90$) (see  generated sample images in Appendix~\ref{fig:sample_cottonweeds_stylegan3}) due to advanced improvements, such as fourier features and $1 \times 1$ convolution kernels, that are applied on the generator network of StyleGAN2.

\begin{table}[!ht]
\centering
\renewcommand{\arraystretch}{1.0}

\resizebox{0.4\textwidth}{!}{
\begin{tabular}{lcccc}
\hline
\multirow{2}{*}{Models} & \multicolumn{4}{c}{CottonWeedID15}   \\ \cline{2-5} 
                        & \multicolumn{1}{c}{FID}            & \multicolumn{1}{c}{IS}           & \multicolumn{1}{c}{Prc}      & Rec    \\ \hline \hline
BigGAN                  & \multicolumn{1}{c}{107.22}              & \multicolumn{1}{c}{5.38}            & \multicolumn{1}{c}{0.5605}               & 0.4315              \\ 
StyleGAN2                 & \multicolumn{1}{c}{43.90}              & \multicolumn{1}{c}{\textbf{7.39}}           & \multicolumn{1}{c}{\textbf{0.6850}}               & 0.4430                 \\ 
StyleGAN3                    & \multicolumn{1}{c}{26.28}              & \multicolumn{1}{c}{5.81}            & \multicolumn{1}{c}{0.5555}               & 0.7975           \\ \hline
ADM                     & \multicolumn{1}{c}{25.08}          & \multicolumn{1}{c}{6.08}         & \multicolumn{1}{c}{0.6550}   & \textbf{0.8085}   \\ 
ADM-DDIM                & \multicolumn{1}{c}{40.76}          & \multicolumn{1}{c}{5.05}         & \multicolumn{1}{c}{0.6130}           & 0.7300                 \\ 
ADM-G                   & \multicolumn{1}{c}{\textbf{24.14}} & \multicolumn{1}{c}{6.70} & \multicolumn{1}{c}{0.6805} & 0.8035     \\ \hline
\end{tabular}}
\caption{Performance comparison between the diffusion models and state-of-the-art GANs. Best values are in bold.}
\label{tab:benchmarks}
\end{table}

\begin{table*}[!ht]
\centering
\renewcommand{\arraystretch}{1.1}
\resizebox{0.55\textwidth}{!}{
\begin{tabular}{llccc}
\hline
\multirow{2}{*}{Index} & \multirow{2}{*}{Models} & \multicolumn{3}{c}{CottonWeedID15} \\ \cline{3-5} 
                       &                         & \multicolumn{1}{c}{Top-1 acc. (\%)} & \multicolumn{1}{c}{Precision (\%)} & Recall (\%)  \\ \hline \hline
\multirow{4}{*}{Original}     & VGG16                   & \multicolumn{1}{c}{98.02 ± 0.25}                   & \multicolumn{1}{c}{97.66 ± 0.19}              & 97.44 ± 0.36                   \\ 
                       & Inception-v3            & \multicolumn{1}{c}{97.30 ± 0.48}                   & \multicolumn{1}{c}{96.74 ± 0.86}              & 96.06 ± 0.96          \\  
                       & DenseNet121                & \multicolumn{1}{c}{98.92 ± 0.37}                   & \multicolumn{1}{c}{98.48 ± 0.46}              & 98.44 ± 0.67       \\  
                       & ResNet50            & \multicolumn{1}{c}{98.43 ± 0.39}                   & \multicolumn{1}{c}{97.92 ± 0.47}              & 97.96 ± 0.52          \\ \hline
\multirow{4}{*}{With augmentation}     & VGG16   & \multicolumn{1}{c}{98.25 ± 0.82}                  & \multicolumn{1}{c}{97.65 ± 1.15}              & 98.19 ± 1.07 \\  
                       & Inception-v3            & \multicolumn{1}{c}{98.47 ± 0.33}                   & \multicolumn{1}{c}{97.63 ± 0.54}              & 98.27 ± 0.51      \\ 
                       &  DenseNet121              & \multicolumn{1}{c}{99.28 ± 0.29}                   & \multicolumn{1}{c}{98.83 ± 0.46}              & 99.31 ± 0.28         \\ 
                       &ResNet50            & \multicolumn{1}{c}{\textbf{99.37 ± 0.37}}                   & \multicolumn{1}{c}{\textbf{99.13 ± 0.53}}              & \textbf{99.44 ± 0.43}    \\ \hline
      \multicolumn{2}{l}{Maximum improvements}                        & \multicolumn{1}{c}{1.20 \%}                  & \multicolumn{1}{c}{1.24\%}              & 2.30\% \\ \hline
\end{tabular}}
\caption{Performance comparison (mean ± standard deviation) of DL models on weed classification trained with and without the samples generated with the proposed data augmentation. Best values are in bold.}
\label{tab:classifiers}
\end{table*}

\begin{table*}[!ht]
\centering
\renewcommand{\arraystretch}{1.2}
\resizebox{0.65\textwidth}{!}{
\begin{tabular}{lccccc}
\hline
       \multirow{2}{*}{Models} & \multicolumn{5}{c}{Proportion of real data in the training sets}                                                     \\ \cline{2-6} 
                                &                \multicolumn{1}{c}{10\%} & \multicolumn{1}{c}{40\%} & \multicolumn{1}{c}{60\%} & \multicolumn{1}{c}{80\%} & 100\% \\ \hline \hline
 VGG16         &      \multicolumn{1}{c}{95.10 ± 0.70}     & \multicolumn{1}{c}{96.40 ± 0.62}     & \multicolumn{1}{c}{97.75 ± 0.86}     & \multicolumn{1}{c}{97.17 ± 0.26}     &   98.02 ± 0.25    \\ 
                                 Inception-v3   & \multicolumn{1}{c}{94.02 ± 0.34}     & \multicolumn{1}{c}{95.73 ± 0.55}     &  \multicolumn{1}{c}{96.22 ± 0.88}     & \multicolumn{1}{c}{96.45 ± 0.66}     &   97.30 ± 0.48    \\  
                                 DenseNet121  & \multicolumn{1}{c}{96.24 ± 0.72}     & \multicolumn{1}{c}{97.75 ± 0.69}     & \multicolumn{1}{c}{98.43 ± 0.57}     & \multicolumn{1}{c}{98.70 ± 0.40}     &   98.92 ± 0.37  \\ 
                                 ResNet50  & \multicolumn{1}{c}{96.67 ± 0.54}     & \multicolumn{1}{c}{97.98 ± 0.45}     & \multicolumn{1}{c}{98.65 ± 0.69}     & \multicolumn{1}{c}{98.83 ± 0.33}     &   98.43 ± 0.39  \\ \hline
\end{tabular}}
\caption{Testing accuracy (\%) (mean ± standard deviation) of four DL models on the CottonWeedID15 dataset with different proportions of real data in the training sets.}
\label{tab:data_pro}
\end{table*}

\subsection{Performance of weed recognition algorithms}\label{sec:classifiers}
After training, we generate 5,000 weed images and apply the Realism score-based cleanup approach (Section~\ref{sec:realism}) to remove the lower-quality samples (see Appendix~\ref{fig:failure}), resulting in about a reduction of 4,000 weed samples with 400 images for each weed class. The generated samples are then added to the training set and partitioned into five subsets according to the Monte Carlo cross-validation approach \citep{raschka2018model}. The new training sets are partitioned to the training (85\%) and validation (15\%)  sets to train the DL models. The testing performance is evaluated on the hold-out testing set (Section~\ref{sec:cottonweedid15}) to ensure fair and reliable evaluation and evaluate how well that DL models perform on unseen data.

Table~\ref{tab:classifiers} shows the testing performance of four DL models, VGG16 \citep{simonyan2014very}, Inception-v3 \citep{szegedy2016rethinking}, DenseNet161 \citep{huang2017densely}, and ResNet50 \citep{he2016deep}, for weed classification on CottonWeedID15 dataset with (i.e., with augmentation) and without data augmentation (i.e., original). It is evident that considerable improvements are achieved for all DL models by incorporating an expanded datasets with synthetic images, in terms of faster converge speed (see training curves in Appendix~\ref{fig:train_curves}) and higher accuracy and lower losses. For example, with the augmented images, Inception-v3 and ResNet-50 yield classification accuracies of 98.47\% and 99.37\% on the CottonWeedID15 dataset, respectively, representing about 1.20\% and 0.95\% improvements over the baseline models without image augmentation. With the augmented dataset, precision and recall are also observed with apparent improvements, showing 1.24\% and 2.30\% maximum improvements, respectively.

Table~\ref{tab:data_pro} shows the testing performance (Top-1 accuracy) of the DL models on CottonWeedID15 dataset trained with different ratios of real images and synthetic images. All DL models achieve an overall classification accuracy of over 94\% even when only 10\% real samples (i.e., 90\% synthetic data) are used for training, which is comparable with the results obtained by classifiers trained with  all real samples (the 100\% column), demonstrating the high quality of the generated samples. It is interesting to note that ResNet50 is able to achieve higher testing accuracies with 60\% or 80\% real samples in the training sets than with all real images on CottonWeedID15 dataset. This may be due to that the diffusion models can cover the data distribution of the training sets and also learn features from different weed classes, resulting in highly diverse samples.

%%%%%%%%%%%%%%%%%%%%%%%%%%%%%%%%%%%%%%%%%%%%%%%%%%%%%%%%%%%%%%%%%%%%%%%%%%%%%%%%%%%%%%
\section{Conclusion and Future Works}
\label{sec:conclu}
In this paper, we presented the first study on adopting diffusion models to generate high-quality synthetic images for weed recognition with comprehensive evaluations on a common weed dataset. Furthermore, a novel realism score-based approach was developed to distill the generated images to remove low-quality samples. Comprehensive experiments showed that the proposed approach consistently outperformed the state-of-the-art GANs in terms of $FID$. Furthermore, pronounced performance improvements on weed classification  were observed for all four DL models if trained on the datasets augmented with generated synthetic weed images. 

In our future work, we plan to further explore the application of diffusion models to generating higher-resolution images (e.g., $512 \times 512$ and $1024 \times 1024$) for agricultural applications by exploiting techniques such as upsampling \citep{dhariwal2021diffusion} and cascaded diffusion models \citep{ho2022cascaded}. In addition, since the current sampling process heavily relies on the class classifier guidance to generate high-quality images while latest classifier-free diffusion models \citep{ho2022classifier} suggest that non-classifier diffusion models are also capable of achieving comparable performance as classifier-guided approaches, we will explore classifier-free diffusion models in our future work to facilitate the implementation. Combing GANs and diffusion models \citep{xiao2021tackling, wang2022diffusion} to generate realistic images with stabilized training would also be a potential research field.

\section*{Authorship Contribution}
\textbf{Dong Chen}: Formal analysis, Software, Writing - original draft;
\textbf{Xinda Qi}: Software, Writing - review  \& editing;
\textbf{Yu Zheng}: Resources, Writing - review  \& editing;
\textbf{Yuzhen Lu}: Writing - review \& editing; 
\textbf{Zhaojian Li}: Supervision, Resources, Writing - review \& editing;   

% \section*{Acknowledgement}
% This work was supported in part by Cotton Incorporated award 21-005. The authors thank Mr. Bailley Terrel for helping annotate the weed images and Dr. Te-Ming Paul Tseng for the assistance in weed identification. 

\typeout{}
\bibliography{ref}

\onecolumn
\appendix
\section{Classifier Scales}\label{app:classifier_scales}
\begin{figure*}[!ht]
  \centering
  \includegraphics[width=0.99\textwidth]{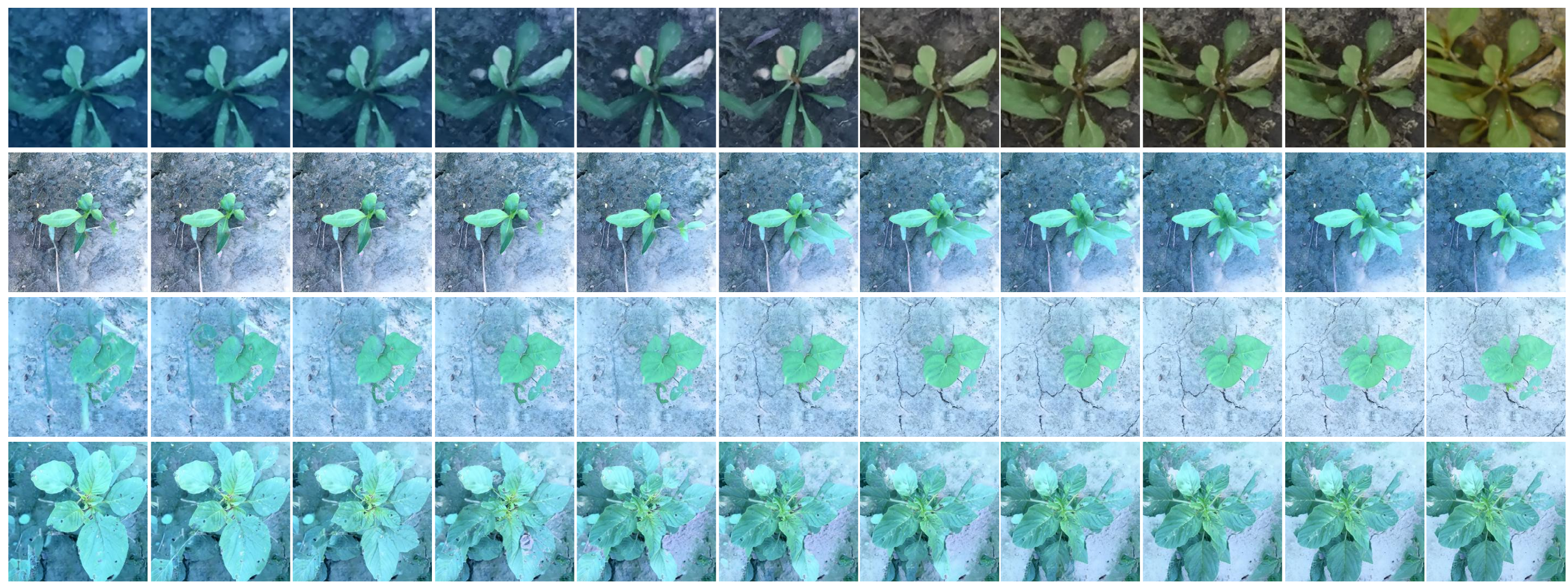}
  \caption{Generated sample images using different classifier scales (evenly ranging from 0 to 5) for the CottonWeedID15 dataset.}
  \label{fig:classifier_scales}
\end{figure*}

Figure~\ref{fig:classifier_scales} shows the sample images generated with different classifier scales (evenly increasing from 0 to 5.0). Higher classifier scales help the model generate higher-fidelity images.

\section{Generated Image Samples}\label{app:image_samples}
\begin{figure*}[!ht]
  \centering
  \includegraphics[width=0.99\textwidth]{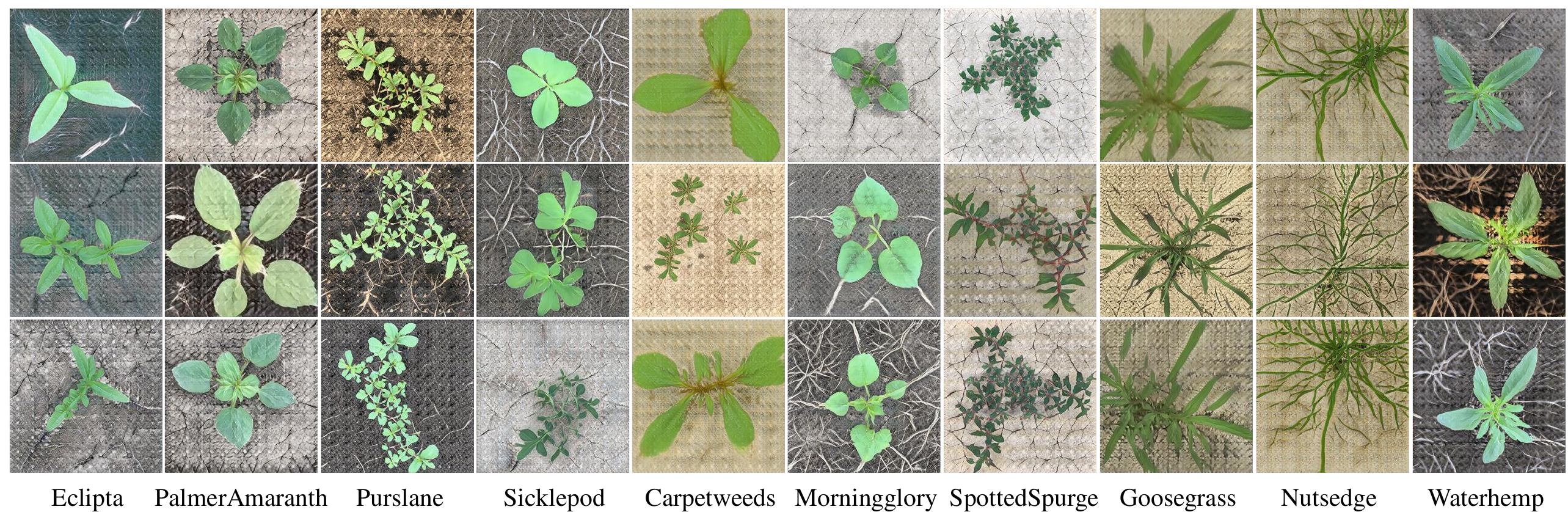}
  \caption{Generated weed samples from the CottonWeedID15 dataset by the BigGAN  \citep{brock2018large} model. Each column represents one weed class.}
  \label{fig:sample_cottonweeds_biggan}
\end{figure*}

Figure~\ref{fig:sample_cottonweeds_biggan} shows the generated weed samples from the CottonWeedID15 dataset by the BigGAN \citep{brock2018large} model. The image quality is relatively low due to the grid effects \citep{odena2016deconvolution, zhang2019gan}. 

\begin{figure*}[!ht]
  \centering
  \includegraphics[width=0.99\textwidth]{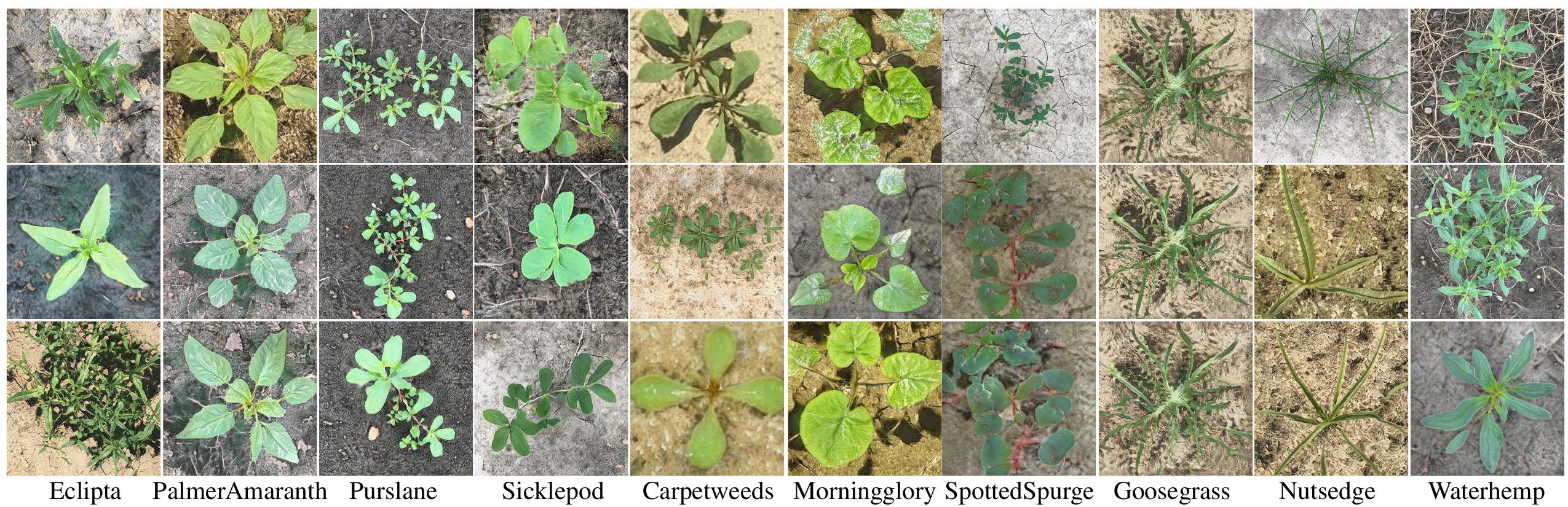}
  \caption{Generated weed samples from the CottonWeedID15 dataset by the StyleGAN2 model. Each column represents one weed class.}
  \label{fig:sample_cottonweeds_stylegan2}
\end{figure*}

% BigGAN \citep{brock2018large}, StyleGAN2 \citep{karras2020analyzing}, and StyleGAN3 \citep{karras2021alias}
Figure~\ref{fig:sample_cottonweeds_stylegan2} shows the generated weed samples from the CottonWeedID15 dataset by StyleGAN2 \citep{karras2020analyzing} model. StyleGAN2 is able to generate high-fidelity weed images but with less diversity.

\begin{figure*}[!ht]
  \centering
  \includegraphics[width=0.99\textwidth]{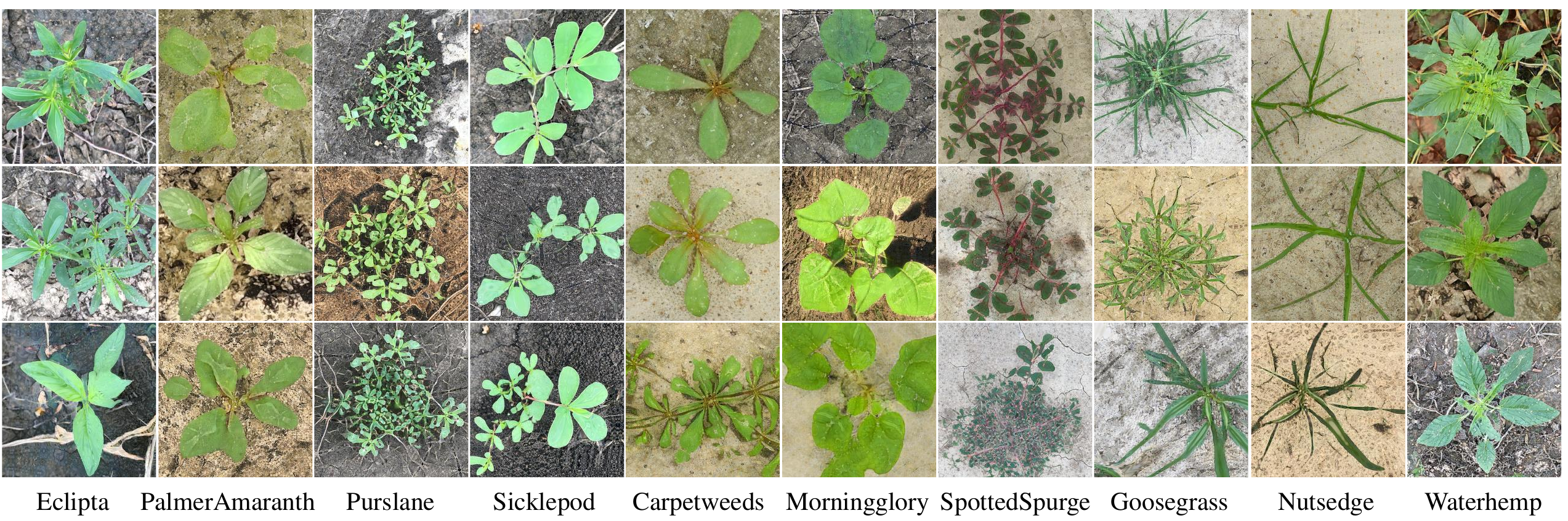}
  \caption{Generated weed samples from the CottonWeedID15 dataset by the StyleGAN3 model. Each column represents one weed class.}
  \label{fig:sample_cottonweeds_stylegan3}
\end{figure*}

Figure~\ref{fig:sample_cottonweeds_stylegan3} shows the generated weed samples from the CottonWeedID15 dataset by the StyleGAN3 \citep{karras2021alias} model. StyleGAN3 is able to generate high-fidelity weed images and outperforms StyleGAN2 in terms of higher FID (see Section~\ref{sec:baselines}).

\section{Lower-quality Samples}\label{app:failure}
\begin{figure*}[!ht]
  \centering
  \includegraphics[width=0.99\textwidth]{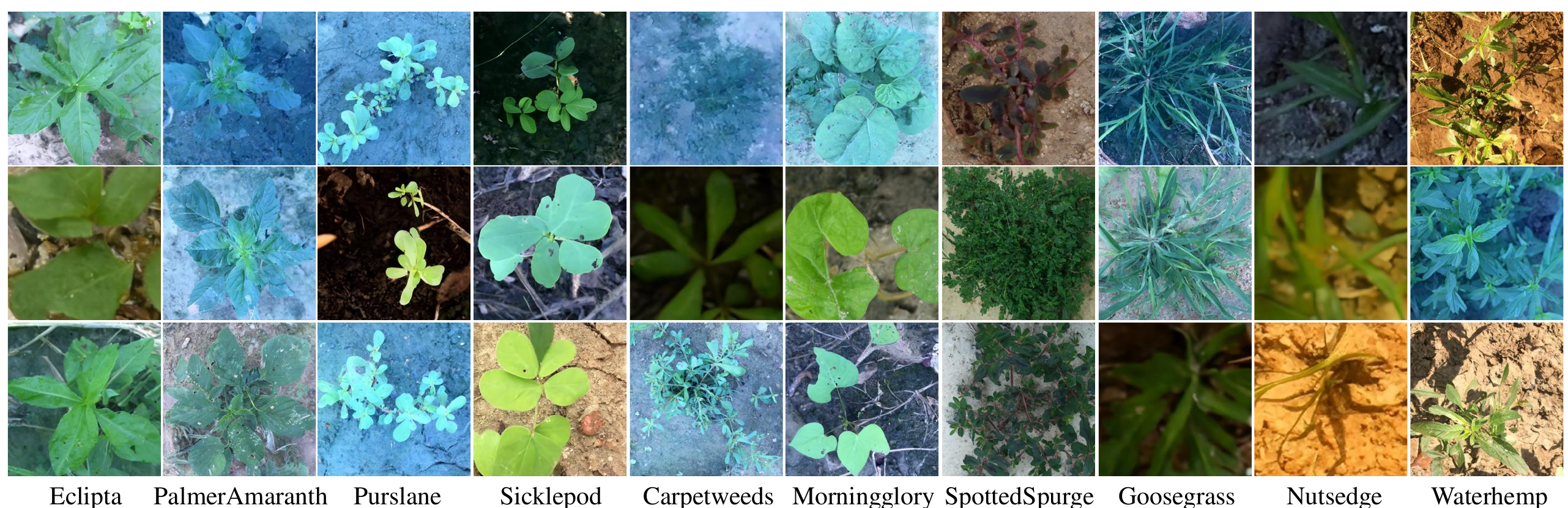}
  \caption{Low-quality samples generated by the diffusion model and filtered by our Realism score-based approach. Each column represents one weed class.}
  \label{fig:failure}
\end{figure*}

Figure~\ref{fig:failure} shows the low-quality weed samples generated by the diffusion model and later removed by the Realism score-based approach (Section~\ref{sec:realism}). Most of the low-quality samples have poor image resolution and color contrast.

\section{Training Curves}\label{app:training_curves}
\begin{figure*}[!ht]
    \centering
    \subfloat{{\includegraphics[width=.472\linewidth]{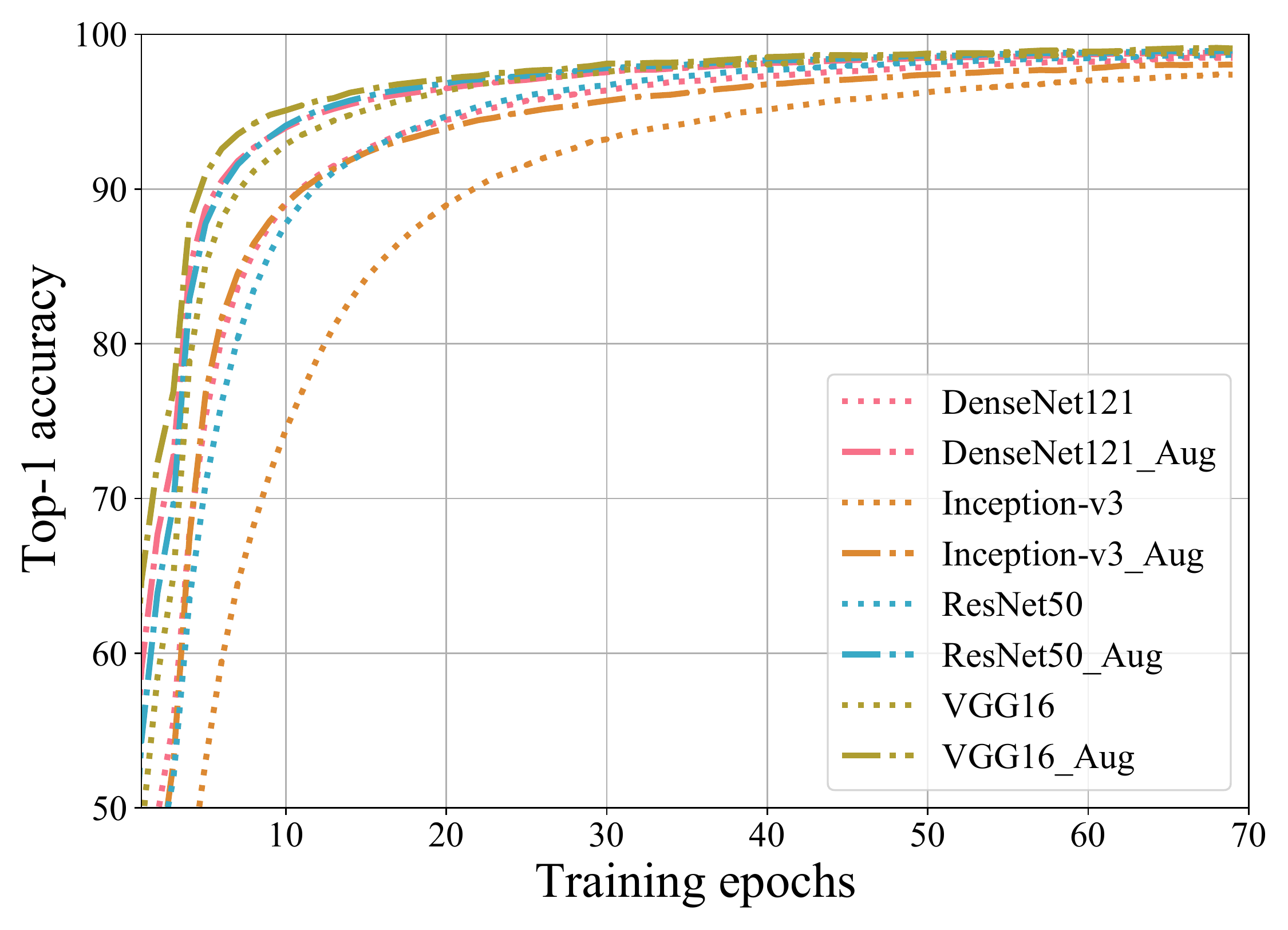} }}%
    \qquad
    \subfloat{{\includegraphics[width=.472\linewidth]{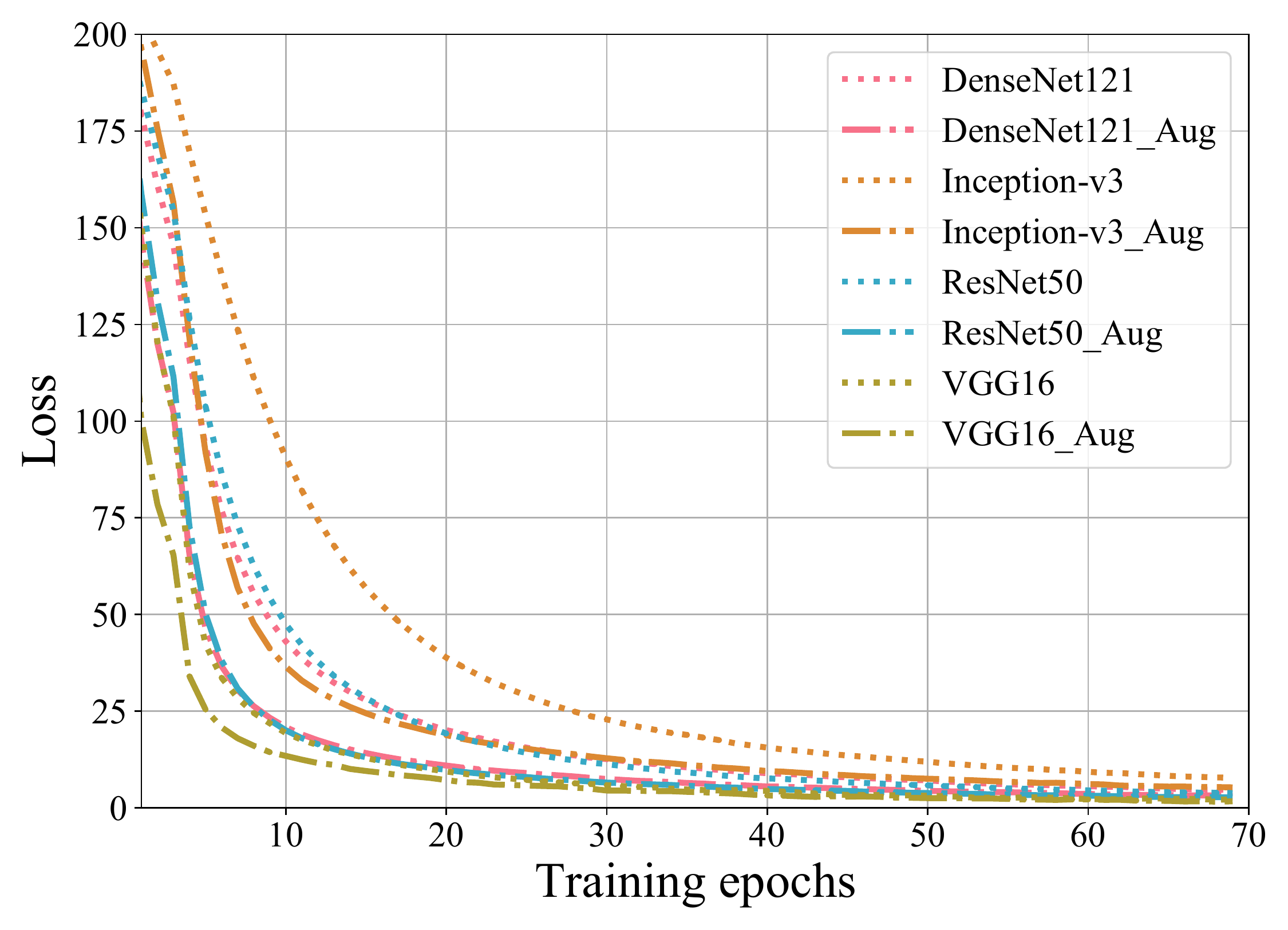}}}
    \caption{Training accuracy and loss for the DL models with and without data augmentation.}
    \label{fig:train_curves}
\end{figure*}

Figure~\ref{fig:train_curves} shows the training curves (accuracy and losses) for the DL models with and without data augmentation. It is clear that training with augmented dataset achieved faster convergence, accuracy, and lower loss.

\end{document}